\documentclass[10pt,twocolumn,letterpaper]{article}

\usepackage[pagenumbers]{cvpr} 

\usepackage{multirow}
\usepackage{arydshln}
\usepackage{amssymb}
\usepackage{pifont}
\usepackage{nicematrix}

\usepackage{lineno}

\definecolor{cvprblue}{rgb}{0.21,0.49,0.74}
\usepackage[pagebackref,breaklinks,colorlinks,allcolors=cvprblue]{hyperref}

\def\paperID{9445} 
\def\confName{CVPR}
\def\confYear{2025}
\definecolor{lightbluee}{rgb}{0.0007, 0.44, 0.737}
\title{Spiking Transformer with Spatial-Temporal Attention}

\author{Donghyun Lee, Yuhang Li, Youngeun Kim, Shiting Xiao, Priyadarshini Panda\\
Department of Electrical Engineering, Yale University\\
{\tt\small \{donghyun.lee, yuhang.li, youngeun.kim, ginny.xiao, priya.panda\}@yale.edu}}

\begin{document}
\maketitle
\begin{abstract}
Spike-based Transformer presents a compelling and energy-efficient alternative to traditional Artificial Neural Network (ANN)-based Transformers, achieving impressive results through sparse binary computations. However, existing spike-based transformers predominantly focus on spatial attention while neglecting crucial temporal dependencies inherent in spike-based processing, leading to suboptimal feature representation and limited performance. To address this limitation, we propose Spiking Transformer with \textbf{S}patial-\textbf{T}emporal \textbf{Atten}tion (\textbf{STAtten}), a simple and straightforward architecture that efficiently integrates both spatial and temporal information in the self-attention mechanism. STAtten introduces a block-wise computation strategy that processes information in spatial-temporal chunks, enabling comprehensive feature capture while maintaining the same computational complexity as previous spatial-only approaches. Our method can be seamlessly integrated into existing spike-based transformers without architectural overhaul. Extensive experiments demonstrate that STAtten significantly improves the performance of existing spike-based transformers across both static and neuromorphic datasets, including CIFAR10/100, ImageNet, CIFAR10-DVS, and N-Caltech101. The code is available at \href{https://github.com/Intelligent-Computing-Lab-Yale/STAtten}{https://github.com/Intelligent-Computing-Lab-Yale/STAtten}.
\end{abstract}    
\section{Introduction}
\label{sec:intro}
Spiking Neural Networks (SNNs) are emerging as a promising alternative to conventional Artificial Neural Networks (ANNs) \cite{maass1997networks,roy2019towards} due to their bio-inspired, energy-efficient computing paradigm. By leveraging sparse binary spike-based computations, SNNs can achieve significant energy savings while enabling deployment across various neuromorphic computing platforms such as TrueNorth \cite{akopyan2015truenorth}, Loihi \cite{davies2018loihi}, and Tianjic \cite{pei2019towards}. However, the challenge lies in effectively processing and learning from these discrete, time-dependent spike patterns while maintaining computational efficiency and accuracy. Traditional SNNs have primarily relied on convolution-based architectures adapted from successful ANN models like VGGNet \cite{wu2019direct,sengupta2019going} and ResNet \cite{zheng2021going,hu2024advancing}. While these architectures benefit from computational efficiency, they often struggle with binary spike activations, leading to information loss and degraded performance. This limitation has driven researchers to explore alternative architectures that can better handle sparse, temporal data while maintaining the energy benefits of spike-based computation.
\begin{figure}
    \centering
    \includegraphics[width=1.0\linewidth]{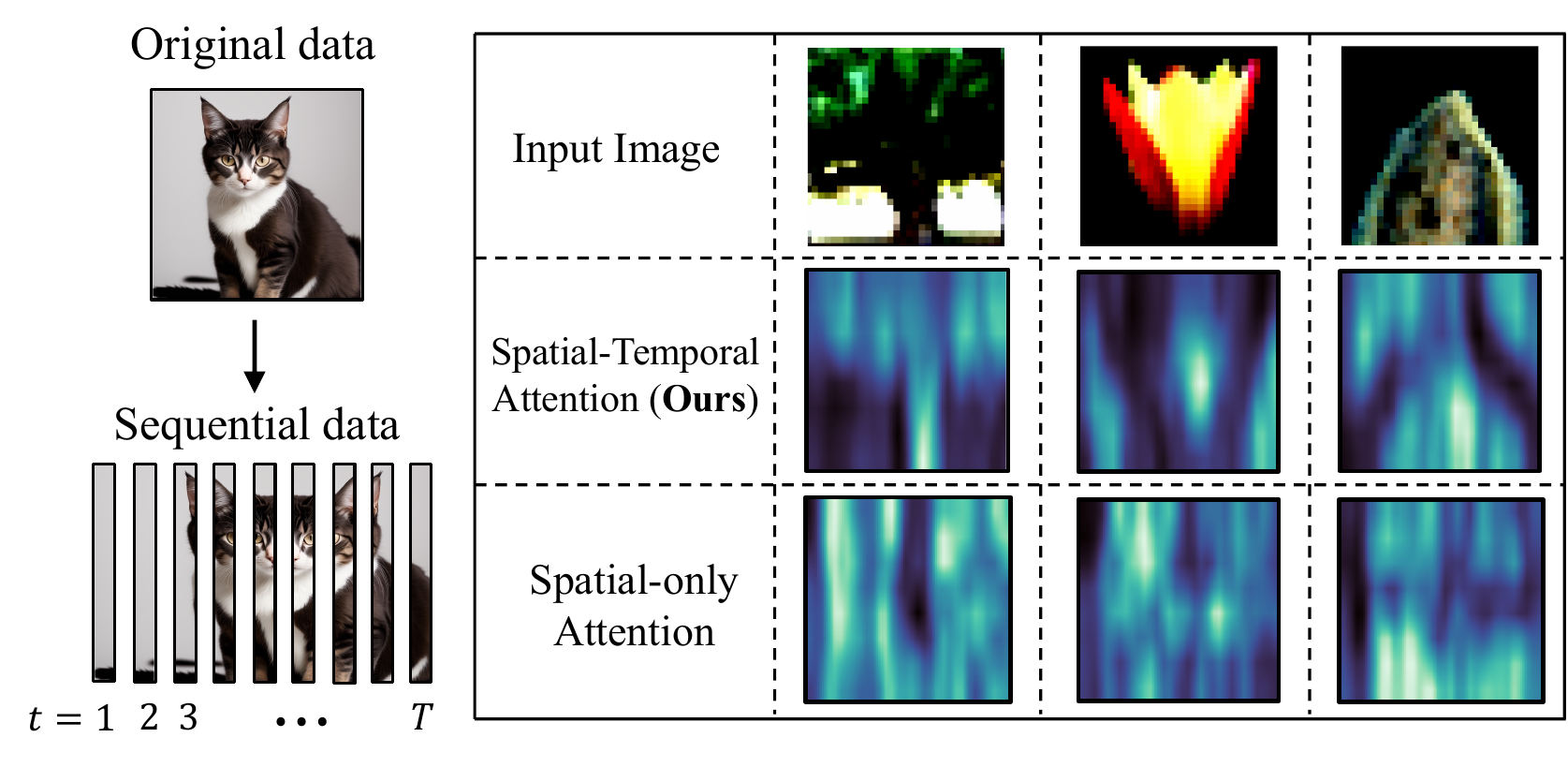}
    \caption{Heatmaps of spatial-only attention versus our STAtten on sequential CIFAR100 dataset. Input images are divided column-wise, where each column corresponds to one timestep.}
    \label{heatmap}
    \vspace{-4mm}
\end{figure}
Transformer architectures, with their remarkable success across various domains \cite{vaswani2017attention, dosovitskiy2020image, arnab2021vivit, chen2021decision}, have emerged as a promising direction for enhancing SNN capabilities. Recent works have introduced spike-formed transformers that adapt the transformer's self-attention mechanism to the spike-based domain for diverse tasks, such as, object tracking \cite{zhang2022spiking}, computer vision \cite{zhou2022spikformer, yao2024spike, yao2024spike2, shi2024spikingresformer, zhou2024qkformer, xiao2025respike}, depth estimation \cite{zhang2022spike}, and speech recognition \cite{tian2020spike, wang2023complex}. These architectures, such as Spikformer \cite{zhou2022spikformer} and Spike-driven Transformer  \cite{yao2024spike}, implement innovative binary representations for query (Q), key (K), and value (V) computations while eliminating the computationally expensive Softmax function. This binary attention design preserves the essential characteristics of spike-based computation while enabling more efficient information processing compared to conventional transformer architectures. However, current spike-formed transformers predominantly rely on spatial-only attention, overlooking the inherently dynamic and temporal nature of spike events. To understand these different attention patterns, we compare spatial-only and spatial-temporal attention on sequential CIFAR100, as shown in Fig.~\ref{heatmap}. The visualization reveals that spatial-only attention focuses solely on vertical (spatial) relationships within each timestep, missing crucial object features that evolve over time. In contrast, our spatial-temporal attention captures both vertical (spatial) and horizontal (temporal) relationships, enabling comprehensive feature representation of sequential inputs.

In this work, we introduce Spiking Transformer with \textbf{S}patial-\textbf{T}emporal \textbf{Atten}tion (\textbf{STAtten}), which integrates both spatial and temporal information within the self-attention mechanism. Through analysis of memory requirements and neuronal activity patterns, we design a block-wise computation strategy with local temporal correlations. By eliminating Softmax operation, we maintain the same computational complexity $(O(TND^2))$ as existing spike-based transformers \cite{zhou2022spikformer, yao2024spike2, shi2024spikingresformer, zhou2024qkformer}. Furthermore, our approach can be flexibly integrated into existing architectures without structural modifications, enhancing their representational capacity. Experiments on both static and dynamic datasets demonstrate consistent performance improvements over spatial-only attention architectures. The main contributions of our work are as follows:
\begin{itemize}
\item We identify limitations in current spiking transformers that rely on spatial-only attention through empirical analysis, demonstrating the importance of capturing temporal dependencies inherent in spike-based processing.

\item We propose STAtten, a block-wise spatial-temporal attention mechanism for spike-based transformers that maintains the original computational complexity $(O(TND^2))$.

\item We introduce a flexible plug-and-play design that enables STAtten to be integrated into existing spike-based transformers without compromising their core architecture.

\item Through extensive experiments across both static (CIFAR10/100, ImageNet) and neuromorphic (CIFAR10-DVS, N-Caltech101) datasets, we demonstrate that STAtten consistently improves performance across different backbone architectures.
\end{itemize}

\section{Related Works}
\label{sec:formatting}
\textbf{Training Strategies for SNNs. }
Significant progress has been made in developing effective training strategies for SNNs, primarily following two approaches: ANN-to-SNN conversion and direct training. The conversion approach transforms pre-trained ANNs into SNNs, leveraging established ANN architectures to enhance SNN performance. Several studies \cite{deng2021optimal, han2020deep, li2021free, diehl2015fast} have formalized this conversion process and demonstrated its effectiveness in transferring ANN knowledge to the spike domain. Direct training through backpropagation, while fundamental to ANNs, presents unique challenges in SNNs due to the non-differentiable nature of spike events. The introduction of surrogate gradients has been crucial in enabling effective backpropagation-based training in SNNs, as demonstrated in pioneering works \cite{shrestha2018slayer, chowdhury2021spatio, wu2019direct}. Notably, direct training with surrogate gradients has shown superior performance compared to conversion methods across various tasks, including image classification \cite{fang2021incorporating, hu2024advancing, yin2024mint,yin2024workload,li2021differentiable}, semantic segmentation \cite{kim2022beyond}, and object detection \cite{su2023deep}, while maintaining the energy efficiency inherent to spike-based computation.

\noindent\textbf{Spiking Transformer. }
Despite the energy efficiency of convolution-based SNNs, their performance often lags behind ANNs due to information loss from sparse binary activations. To address this limitation, researchers have adapted transformer architectures to the SNN domain, leveraging their powerful self-attention mechanisms and ability to capture long-range dependencies. This adaptation has led to several innovative spike-formed transformer architectures. Spikformer \cite{zhou2022spikformer} introduced the first SNN-based transformer architecture, establishing foundational principles for spike-formed self-attention mechanisms. Their key innovation lies in eliminating the Softmax function from self-attention computations, justified by two key observations: (1) the binary nature of spike-based Q, K, and V values, and (2) the redundancy of Softmax scaling for binarized operations. Building upon this foundation, Spike-driven Transformer \cite{yao2024spike} enhanced the architecture with a novel self-attention block incorporating masking and sparse addition to optimize power efficiency. Additionally, inspired by MS-ResNet \cite{hu2024advancing}, Spike-driven Transformer reimagined residual connections to propagate membrane potential rather than spikes. Recent works have further extended these concepts, leading to architectures like SpikingReformer \cite{shi2024spikingresformer}, QKFormer \cite{zhou2024qkformer}, and Spike-driven Transformer-V2 \cite{yao2024spike2}, each introducing unique optimizations for spike-based processing. However, existing spiking transformer architectures primarily focus on spatial attention, neglecting the temporal dynamics inherent in spike-based computation. 


\section{Preliminary}
\noindent\textbf{Leaky Integrate-and-Fire Neuron. }
As a fundamental building block of SNNs, the Leaky Integrate-and-Fire (LIF) neuron \cite{burkitt2006review} has emerged as an important component for energy-efficient computation. The LIF neuron is a non-linear activation function that determines whether neurons fire spikes as follows:
\begin{align}
    {\mathbf{u}[t+1]^l = \tau \mathbf{u}[t]^l+\mathbf{W}^lf(\mathbf{u}[t]^{l-1})} \label{lif1} \\
    f(\mathbf{u}[t]^l) = 
    \begin{cases} 
    1 & \text{if} \; \mathbf{u}[t]^l > V_{th}, \\ 0 & \text{otherwise}
    \end{cases},
\end{align}
\noindent where, $\mathbf{u}[t]^l$ is the membrane potential in $l$-th layer at timestep $t$, $\tau\in(0,1]$ is the leaky factor for membrane potential leakage, $\mathbf{W}^l$ is the weight of $l$-th layer, and $f(\cdot)$ is the LIF function with firing threshold $V_{th}$. Therefore, when the membrane $\mathbf{u}[t]^l$ is higher than $V_{th}$, the LIF function fires a spike and the membrane potential is reset to 0.

\noindent\textbf{Vanilla Self-attention. }
Transformers~\cite{vaswani2017attention, dosovitskiy2020image} have surpassed convolution-based architectures due to their unique self-attention mechanism, which can capture global features through long-range dependency. For a floating-point input tensor $\mathbf{X}_f \in \mathbb{R}^{N \times D}$ with $N$ tokens, and $D$ features, we formulate the self-attention as follows:
\begin{equation}
    \begin{aligned}
     \mathbf{Q}_f = \mathbf{W_Q}\mathbf{X}_f,\quad \mathbf{K}_f = \mathbf{W}_f\mathbf{X}_f, \quad \mathbf{V}_f = \mathbf{W_V}\mathbf{X}_f
    \end{aligned}
\end{equation}
\begin{equation}
    \begin{aligned}
     \mathrm{Attn} = \texttt{Softmax} (\frac{\mathbf{Q}  \mathbf{K}^\top}{\sqrt{D}})\mathbf{V}, \ \ \ \mathbf{Q}_f,\mathbf{K}_f,\mathbf{V}_f \in \mathbb{R}^{N\times D},
    \end{aligned}
\end{equation}
where $\mathbf{W_Q}$, $\mathbf{W_K}$, $\mathbf{W_V} \in \mathbb{R}^{D \times D}$ are linear projections for floating-point $\mathbf{Q}_f$, $\mathbf{K}_f$, $\mathbf{V}_f$, respectively. Note that the computational complexity of attention is $\mathcal{O}(N^2D)$ due to its matrix multiplication operation. 

\noindent\textbf{Spike-based Self-attention. } The first spike-based transformer work, Spikformer~\cite{zhou2022spikformer}, proposes Spiking Self Attention (SSA) with binary $\mathbf{Q}$, $\mathbf{K}$, $\mathbf{V}$. For binary input tensor $\mathbf{X} \in \mathbb{R}^{T \times N \times D}$ with $T$ timestep, we formulate the SSA as follows:
\begin{equation}
\begin{aligned}
     \mathbf{Q}, \mathbf{K},  \mathbf{V} &= \text{LIF}(\mathbf{W_Q}\mathbf{X}), \ \text{LIF}(\mathbf{W_K}\mathbf{X}), \ \text{LIF}( \mathbf{W_K}\mathbf{X}), \\
     &\mathrm{SSA}(\mathbf{Q}\mathbf{K}\mathbf{V})=\text{LIF}(\mathbf{Q}\mathbf{K}^{\top}\mathbf{V}\cdot \alpha)
\end{aligned}
\end{equation}
where $\mathbf{W_Q}$, $\mathbf{W_K}$, $\mathbf{W_V} \in \mathbb{R}^{D \times D}$ are linear projections for binary $\mathbf{Q}$, $\mathbf{K}$, $\mathbf{V}$, respectively, and $\alpha$ denotes a scaling factor.

\section{Methodology}
In this section, we present our analysis of spike-based transformers and provide the details of our proposed methodology. We first establish the necessity of spatial-temporal attention in spike-based transformers through an information-theoretic analysis based on entropy measurements. Building upon these observations, we introduce our STAtten mechanism. 
\begin{figure}[t]
\centering
    \begin{tabular}{@{}c@{\hskip 0.01\linewidth}c@{\hskip 0.01\linewidth}c}
\includegraphics[width=0.45\linewidth]{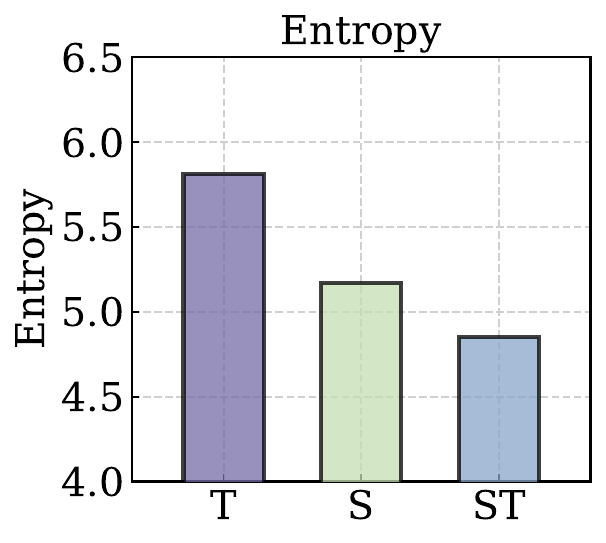} &
\includegraphics[width=0.45\linewidth]{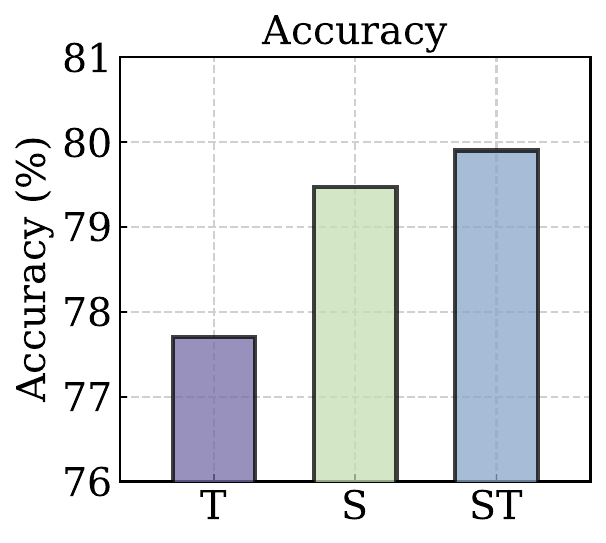}

\\
\end{tabular}
\vspace{-3mm}
\caption{
Comparison between different self-attentions with the CIFAR100 dataset. Analysis of entropy and accuracy for Temporal-only (T), Spatial-only (S), and Spatial-temporal (ST).
} \label{entropy}

  \vspace{-4.5mm}
\end{figure}

\begin{figure}[t]
\centering

    \includegraphics[width=1.0\linewidth]{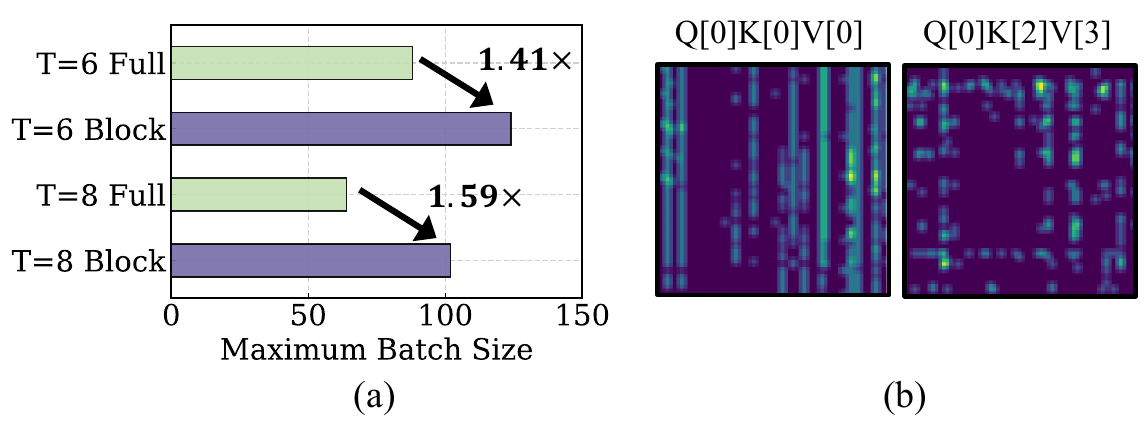}
    \vspace{-6.5mm}
    \caption{(a) Maximum batch size of block-wise STAtten and full spatial-temporal attention (without block partitioning) for running on A5000 GPU with 24GB VRAM memory. (b) Average number of active neurons after QKV computation at different timestep combinations, where [t] indicates timestep index.}
    \label{reason}
\vspace{-3mm}
\end{figure}

\begin{figure*}
    \centering
    \includegraphics[width=17cm]{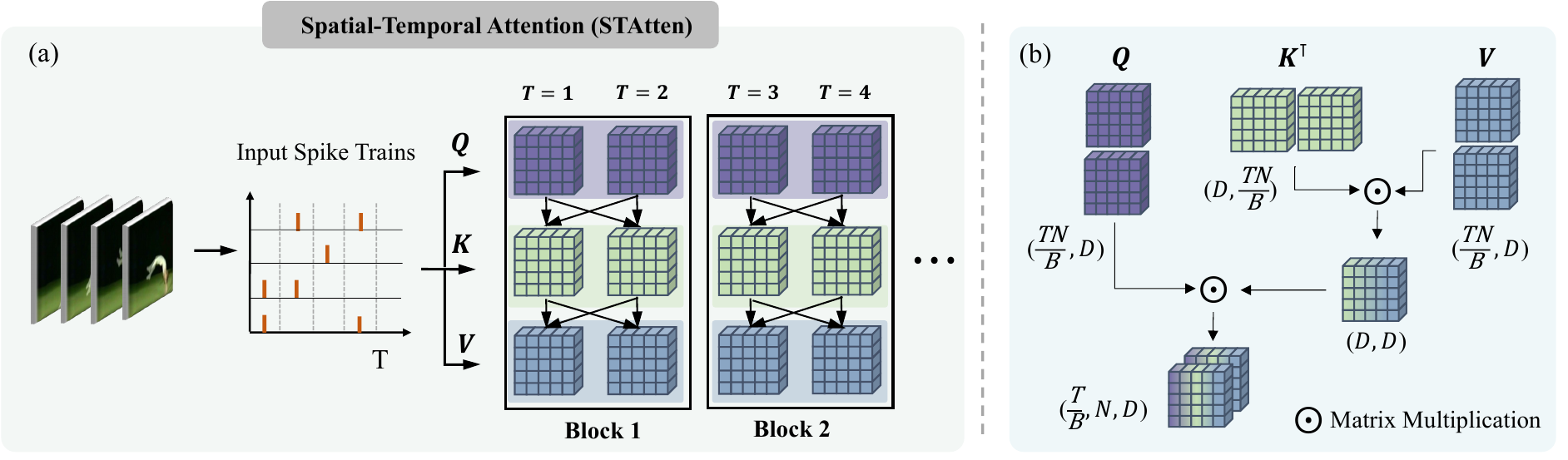}
    \vspace{-4mm}
    \caption{Overview of STAtten architecture. \textbf{(a)} Block-wise temporal attention mechanism. Binary Q, K, V tensors are partitioned into temporal blocks, where black lines indicate paired timestep processing. \textbf{(b)} Computation flow with tensor dimensions, where $T$ is the number of timesteps, $N$ is the number of tokens, $B$ is block size, and $D$ is the feature dimension.}
    \label{method}
\vspace{-3mm}
\end{figure*}

\subsection{Motivation of Spatial-Temporal Attention}
\label{motivation}
 The feature maps of spike-based transformers at each timestep carry distinct information, as their activation distributions are dynamically influenced by both the leaky factor and reset mechanisms. This temporal aspect of information flow raises a fundamental question: how do different attention mechanisms effectively capture and utilize this spatio-temporal information?

First, we define the three different types of attention: Spatial, Temporal, and Spatial-temporal attention with timestep $t$, token position $n$, and feature dimension $d$ as follows:
\begin{equation}
\begin{aligned}
    &\mathrm{S\_Attn}[t,:,:] = \text{LIF}(\mathbf{Q}[t,:,:]\mathbf{K^\top}[t,:,:]\mathbf{V}[t,:,:]\cdot \alpha),\\
    &\mathrm{T\_Attn}[:,n,:] = \text{LIF}(\mathbf{Q}[:,n,:]\mathbf{K^\top}[:,n,:]\mathbf{V}[:,n,:]\cdot \alpha), \\
     &\mathrm{ST\_Attn}[:,:,d] = \text{LIF}(\mathbf{Q}[:,:,d]\mathbf{K}^\top[:,:,d]\mathbf{V}[:,:,d]\cdot \alpha), 
\label{TN}
\end{aligned}
\end{equation}
where $\mathrm{S\_Attn}$, $\mathrm{T\_Attn}$, and $\mathrm{ST\_Attn}$ are spatial, temporal, and spatial-temporal attention, and the $\mathbf{Q}$, $\mathbf{K}$, $\mathbf{V}$  dimensions of $\mathrm{S\_Attn}$, $\mathrm{T\_Attn}$, and $\mathrm{ST\_Attn}$ are $\mathbb{R}^{T \times N \times D}$. The key distinction lies in their attention computation scope: spatial attention operates on token relationships within individual timesteps, temporal attention focuses on cross-timestep feature dependencies, and spatial-temporal attention integrates both spatial and temporal feature correlations. To quantify the information distribution in each attention mechanism, we measure Shannon entropy \cite{gabrie2018entropy} as follows:
\begin{equation}
   H(\mathrm{Attn}) = -\sum_{t=1}^T \sum_{n=1}^N \sum_{d=1}^D \mathbf{\hat p}[t,n,d] \log \mathbf{\hat p}[t,n,d],
\end{equation}
where $\mathbf{\hat p}[t,n,d] = \texttt{Softmax}(\mathrm{Attn}[t,n,d])$. 
Lower entropy values indicate more focused spike patterns, while higher entropy values suggest dispersed patterns~\cite{kim2023exploring}. We analyze this relationship using a pretrained Spike-driven Transformer \cite{yao2024spike} on the CIFAR100 dataset, as shown in Fig.~\ref{entropy}. The results reveal an inverse correlation between entropy and accuracy: temporal-only attention shows high entropy (5.81) but lowest accuracy (77.7\%), while spatial-temporal attention achieves lower entropy (4.85) with higher accuracy (79.9\%). These findings verify that combining spatial and temporal processing leads to more structured feature representations, providing strong motivation for spatial-temporal attention.

However, we argue that fully correlating temporal information across all timesteps like $\mathrm{ST\_Attn}$ in Eq. \ref{TN} introduces two challenges: \textbf{(1)} substantial memory requirements for storing temporal information, and \textbf{(2)} increased dead neurons when correlating distant timesteps due to weak spike similarity. The first challenge arises from matrix multiplication with large temporal dimensions ($TN, D$). To address this, we partition the temporal dimension into blocks, reducing the memory footprint for matrix operations. As shown in Fig. \ref{reason}(a), this block-wise computation strategy leads to around 1.6 $\times$ higher batch size compared to full temporal attention. The second challenge stems from binary matrix multiplication in spike-based processing. Our analysis in Fig. \ref{reason}(b) reveals that QKV computations show higher active neurons when processing same-timestep, while activity decreases with temporal distance.

\subsection{STAtten Mechanism}



Based on these observations, we propose \textbf{S}patial-\textbf{T}emporal \textbf{Atten}tion (\textbf{STAtten}). The block-wise STAtten partitions the temporal sequence into blocks of size $B$, as shown in Fig. \ref{method}(a). We define STAtten as follows: 
\begin{equation}
\begin{aligned}
&\text{STAtten}(\mathbf{X}[b]) = \text{LIF}(\mathbf{Q}[b] \mathbf{K}^\top[b] \mathbf{V}[b] \cdot \alpha), \\
&\text{where } [b] = [iB:(i+1)B,:,d], \quad i \in \{0,1,...,T/B-1\}
\end{aligned}
\end{equation}
where $[iB:(i+1)B,:,d]$ denotes the aggregated features from timesteps $iB$ to $(i+1)B$. For instance, with $T=4$ and $B=2$, features are grouped in pairs of timesteps [0,1], [2,3]. By processing blocks of timesteps together, we capture local temporal relationships while maintaining manageable computation. This approach requires storing only a subset of temporal information at any given time, reducing memory overhead. We set block size $B=2$ with 4 timesteps for static datasets and $B=4$ with 16 timesteps for neuromorphic datasets.


Unlike traditional transformers that require Softmax operations, the binary $\mathbf{Q}$, $\mathbf{K}$, and $\mathbf{V}$ allow flexible reordering of computations.  Generalizing to arbitrary number of timesteps $T$, the computation follows:
\begin{equation}
\begin{aligned}
\mathbf{Q}, \mathbf{K}, \mathbf{V} &\in \mathbb{R}^{\frac{T}{B} \times N \times D} \xrightarrow{\text{reshape}} \mathbf{Q}, \mathbf{K}, \mathbf{V} \in \mathbb{R}^{(\frac{T}{B}N) \times D} \\
\mathbf{K}^\top\mathbf{V} &\in \mathbb{R}^{D \times D} \xrightarrow{\text{attention}} \mathbf{Q}(\mathbf{K}^\top\mathbf{V}) \in \mathbb{R}^{\frac{T}{B} \times N \times D}.
\end{aligned}
\end{equation}
The computation sequence reduces intermediate memory requirements through 
 early $\mathbf{K}^\top\mathbf{V}$ calculation while preserving cross-timestep information flow within blocks. This implementation maintains computational complexity of $\mathcal{O}(TND^2)$, matching spatial-only attention \cite{zhou2022spikformer, yao2024spike, yao2024spike2, zhou2024qkformer, shi2024spikingresformer} while incorporating temporal dependencies. By avoiding the quadratic complexity in timesteps and the number of tokens ($T^2N^2$) that would result from full temporal correlation, our method scales efficiently to longer sequences. Fig. \ref{method}(b) illustrates the complete computation process with corresponding tensor dimensions.

\begin{table}
\caption{Computational complexity and energy consumption of different attention mechanisms. $E_{MAC}$ and $E_{AC}$ are energy costs for MAC and AC operations, $N$, $D$, $T$ denote patches, dimension, timesteps, and $S_Q$, $S_K$, $S_V$ are firing rates of $Q$, $K$, $V$. SDT represents Spike-driven Transformer.}

  \label{table1}
  \centering
  \resizebox{\linewidth}{!}{
  \begin{tabular}{lccc}
    \toprule
     Method & ST & Complexity & Energy   \\
    \midrule
     ViT \cite{dosovitskiy2020image} & \ding{55} & $\mathcal{O}(N^2D)$ &  $E_{MAC} \cdot N^2D$ \\
     ViViT \cite{arnab2021vivit} & \ding{51} & $\mathcal{O}(T^2N^2D)$ &  $E_{MAC} \cdot T^2N^2D$ \\
     \cdashline{1-4}\\[-1.75ex]
     Spikformer \cite{zhou2022spikformer}& \ding{55} & $\mathcal{O}(TND^2)$ &  $E_{AC} \cdot TND^2 \cdot (S_Q + S_K + S_V)$       \\
     SDT \cite{yao2024spike}& \ding{55} & $\mathcal{O}(TND)$ &  $E_{AC} \cdot TND \cdot (S_Q + S_K)$       \\ 
     SDT-V2 \cite{yao2024spike2}& \ding{55} & $\mathcal{O}(TND^2)$ &  $E_{AC} \cdot TND^2 \cdot (S_Q + S_K + S_V)$       \\ 
     QKFormer \cite{zhou2024qkformer}& \ding{55} & $\mathcal{O}(TND^2)$ &  $E_{AC} \cdot TND^2 \cdot (S_Q + S_K + S_V)$       \\

     \cdashline{1-4}\\[-1.75ex]
     STAtten& \ding{51} & $\mathcal{O}(TND^2)$ &  $E_{AC} \cdot TND^2 \cdot (S_Q + S_K + S_V)$       \\ 
     
    \bottomrule
  \end{tabular}}
\end{table}

\begin{table}
\centering
\caption{Performance comparison on sequential CIFAR10/100 datasets, demonstrating the effectiveness of spatial-temporal attention in capturing long-term dependencies.}
\label{seq_tab}
\resizebox{\linewidth}{!}{%
\begin{tabular}[width=0.85\textwidth]{lcc}
    \toprule
    Method & s-CIFAR10 (\%) & s-CIFAR100 (\%)    \\
    \midrule
    PLIF \cite{fang2021incorporating} & 81.47 & 53.38 \\
     \cdashline{1-3}\\[-1.75ex]
     Spikformer \cite{zhou2022spikformer} & 79.29 &  57.17 \\
     \cellcolor{gray!25}STAtten + \cite{zhou2022spikformer}& \cellcolor{gray!25}\textbf{82.45} & \cellcolor{gray!25}\textbf{58.75}   \\
     \cdashline{1-3}\\[-1.75ex]
     Spike-driven Transformer \cite{yao2024spike} & 80.32 &  61.08 \\
     \cellcolor{gray!25}STAtten + \cite{yao2024spike}& \cellcolor{gray!25}\textbf{83.41} & \cellcolor{gray!25}\textbf{64.30}     \\ 
     \cdashline{1-3}\\[-1.75ex]
     SpikingReformer \cite{shi2024spikingresformer}  & 79.25 &  57.48 \\
     \cellcolor{gray!25}STAtten + \cite{shi2024spikingresformer}& \cellcolor{gray!25}\textbf{81.84} & \cellcolor{gray!25}\textbf{58.30}     \\ 

    \bottomrule
    \end{tabular}}
\end{table}

\begin{table*}[t]
   \footnotesize
  \caption{Performance comparison between our methods and previous works on CIFAR10 and CIFAR100 datasets. In the architecture column, Architecture-$L$-$D$ represents $L$ number of encoder blocks and $D$ hidden dimensions. All baseline architectures of spike-based transformer and corresponding STAtten implementations are trained from scratch for fair comparison. }
  \label{cifar_result}
  \centering
  \begin{tabular}{llcccc}
    \toprule
    Method & Architecture  & Timestep & CIFAR10 ($\%$) & CIFAR100 ($\%$)      \\
    \midrule
     tdBN \cite{zheng2021going} & ResNet19 & 4 & 92.92 & - \\
     PLIF \cite{fang2021incorporating} & ConvNet & 8 & 93.50 & - \\
     Dspike \cite{li2021differentiable} & ResNet18 & 6 & 94.25 & 74.24 \\
     DSR \cite{meng2022training} & ResNet18 & 20 & 95.40 & 78.50 \\
     DIET-SNN \cite{rathi2020diet} & VGG16 & 5 & 92.70 & 69.67\\
     SNASNet \cite{kim2022neural} & ConvNet & 5 & 93.64 & 73.04 \\
     \midrule
     
     Spikformer \cite{zhou2022spikformer} & Spikformer-4-384 & 4 & 93.99 & 75.06 \\
     \cellcolor{gray!25}\textbf{STAtten} + \cite{zhou2022spikformer} &  \cellcolor{gray!25}Spikformer-4-384 & \cellcolor{gray!25}4 & \cellcolor{gray!25}\textbf{94.36} & \cellcolor{gray!25}\textbf{75.85} \\
     
     \cdashline{1-5}\\[-1.75ex]
     
     Spike-driven Transformer \cite{yao2024spike}  & Spike-driven Transformer-2-512 & 4 & 95.60 & 78.40 \\
    \cellcolor{gray!25}\textbf{STAtten} + \cite{yao2024spike}  &  \cellcolor{gray!25}Spike-driven Transformer-2-512 & \cellcolor{gray!25}\cellcolor{gray!25}4 & \cellcolor{gray!25}\textbf{96.03} & \cellcolor{gray!25}\textbf{79.85} \\

    \cdashline{1-5}\\[-1.75ex]
     
     SpikingReformer \cite{shi2024spikingresformer} & SpikingReformer-6-384 & 4 & 95.03 & 77.16 \\
    \cellcolor{gray!25}\textbf{STAtten} + \cite{shi2024spikingresformer}&  \cellcolor{gray!25}SpikingReformer-6-384 & \cellcolor{gray!25}4 & \cellcolor{gray!25}\textbf{95.26} & \cellcolor{gray!25}\textbf{77.90} \\
    \cdashline{1-5}\\[-1.75ex]
     
     QKFormer \cite{zhou2024qkformer} & QKFormer-4-384 & 4 & 95.12 & 79.79 \\
    \cellcolor{gray!25}\textbf{STAtten} + \cite{zhou2024qkformer}&  \cellcolor{gray!25}QKFormer-4-384 & \cellcolor{gray!25}4 & \cellcolor{gray!25}\textbf{95.35} & \cellcolor{gray!25}\textbf{80.20} \\
    \bottomrule
  \end{tabular} \\
 \vspace{-2mm}
\end{table*}

\subsection{STAtten with Existing Spiking Transformers}

STAtten is designed to be a plug-and-play module that can enhance various spike-based transformer architectures, including Spikformer \cite{zhou2022spikformer}, Spike-driven Transformer \cite{yao2024spike}, Spike-driven Transformer-V2 \cite{yao2024spike2}, SpikingReformer \cite{shi2024spikingresformer}, and QKformer \cite{zhou2024qkformer} while preserving their unique characteristics.


The key to successful integration lies in preserving each architecture's distinct residual connection strategies while replacing their spatial-only attentions with STAtten. Two primary residual connection mechanisms are prevalent in current architectures: Spike Element-Wise (SEW) shortcut \cite{fang2021deep} and Membrane-Shortcut (MS) \cite{hu2024advancing}. In architectures employing SEW shortcuts, such as Spikformer and QKFormer, spike activations are directly added to the attention output. In contrast, Spike-driven Transformer, Spike-driven Transformer-V2, and SpikingReformer utilize MS-shortcuts that propagate membrane potentials rather than spikes Our STAtten module maintains compatibility with both connection types. We detail the existing spike-based attention mechanisms of each architecture and our integration strategy.



\noindent\textbf{Spikformer \cite{zhou2022spikformer}: } Spikformer proposes Spiking Self Attention (SSA) module which is inspired by the vanilla self-attention of ViT \cite{dosovitskiy2020image} as follows:
\begin{equation}
    SSA(\mathbf{Q}, \mathbf{K}, \mathbf{V}) = \text{LIF}(\mathbf{Q} \odot \mathbf{K}^{\top} \odot \mathbf{V}\cdot \alpha),
\end{equation}
where $\odot$ is matrix multiplication. This work establishes two key principles for spike-based attention: non-softmax operation and flexible $\mathbf{Q}$, $\mathbf{K}$, $\mathbf{V}$ operation ordering. While these principles form the foundation for subsequent spike-based architectures, the SSA module processes features independently at each timestep. We address this limitation by replacing the attention mechanism with STAtten.

\noindent\textbf{Spike-driven Transformer \cite{yao2024spike}: } Spike-driven Transformer introduces Spike-Driven Self-Attention (SDSA), which reformulates attention computation as:
\begin{equation}
   SDSA(\mathbf{Q}, \mathbf{K}, \mathbf{V}) = \mathbf{Q} \otimes \text{LIF}(\sum_c(\mathbf{K} \otimes \mathbf{V})),
\end{equation}
where $\otimes$ denotes element-wise multiplication and $\sum_c$ represents column-wise summation. While SDSA achieves computational efficiency through element-wise operations, this design limits feature correlation even within the spatial domain due to the element-wise multiplication. By integrating STAtten into this architecture, we enable comprehensive spatial-temporal feature correlation.

\noindent\textbf{SpikingReformer \cite{zhou2023spikingformer}: } SpikingReformer identifies two key challenges in adapting conventional $\mathbf{Q}$, $\mathbf{K}$, $\mathbf{V}$ attention to spike-based transformers: (1) floating-point matrix multiplication between $\mathbf{Q}$, $\mathbf{K}$, $\mathbf{V}$ and (2) Softmax operations involving exponentiation and division. To address these challenges, the authors propose Dual Spike Self-Attention (DSSA):
\begin{equation}
  DSSA(\mathbf{X}) = \text{LIF}(\mathbf{X} \odot \mathbf{W}^{\top} \odot \mathbf{X}^{\top}\cdot\alpha),
\end{equation}
where $\mathbf{X}$ and $\mathbf{W}$ denote binary input and linear projection, respectively. We try to verify that these concerns can be addressed by replacing DSSA with STAtten, showing that $\mathbf{Q}$, $\mathbf{K}$, $\mathbf{V}$ matrix multiplication is still compatible with spike-based processing when properly designed.

\noindent\textbf{QKFormer \cite{zhou2024qkformer}: } QKFormer implements dual attention mechanisms: Q-K Channel Attention (QKTA) and SSA. While SSA follows the Spikformer design, QKTA introduces a token-wise attention:
\begin{equation}
  QKTA= \text{LIF}(\sum_N(\mathbf{Q}))\otimes \mathbf{K},
\end{equation}
where $\sum_N$ represents token-wise summation. QKFormer redesigns the architecture by incorporating embedding layers in each encoder. In this architecture, we only replace SSA with STAtten as SSA is the primary mechanism responsible for capturing feature relationships.


As we only modify the self-attention module, the original information flow through residual connections remains unchanged, maintaining the architecture's core characteristics. After computing STAtten across temporal blocks, the features are concatenated and processed through subsequent layers: $\mathbf{F}_{\text{out}} = \text{Concat}({\text{STAtten}(\mathbf{X}{[iB:(i+1)B]})}_{i=0}^{T/B-1})$. This concatenated feature then flows through convolutional and batch normalization layers: $\mathbf{Z} = \text{BN}(\text{Conv}(\mathbf{F}_{\text{out}}))$.

\subsection{Complexity and Energy of Self-attention}
\label{energy}

We analyze both computational complexity and theoretical energy consumption of our proposed architecture, focusing specifically on the self-attention mechanism. A comprehensive evaluation of the overall architecture's energy profile is provided in Supplementary Material. 

As shown in Table \ref{table1}, despite incorporating spatial-temporal attention capabilities, our STAtten maintains the same computational complexity $\mathcal{O}(TND^2)$ as previous spatial-only spiking transformers. This efficiency is achieved through our non-Softmax design, which enables flexible reordering of $\mathbf{Q}$, $\mathbf{K}$, $\mathbf{V}$ operations. In contrast, conventional ANN-based spatial-temporal attention methods like ViViT \cite{arnab2021vivit} require substantially higher complexity ($\mathcal{O}(T^2N^2D)$) due to their softmax-constrained computation order. For energy consumption estimation, we follow the floating-point operations (FLOPs) implementation in 45nm technology \cite{horowitz20141}. The energy costs are categorized into $E_{MAC}$ for multiply-accumulate operations corresponding to 32-/8-bit ANN computations, and $E_{AC}$ for accumulate operations used in binary SNN calculations. The final energy consumption is modulated by the firing rates ($S_Q$, $S_K$, and $S_V$) of the respective Q, K, and V. A detailed analysis of these energy measurements and their implications is presented in Section \ref{main_energy}.

\section{Experiments}

We verify our proposed methods on both static and dynamic datasets, \eg, CIFAR10/100 \cite{krizhevsky2009learning}, ImageNet \cite{deng2009imagenet}, CIFAR10-DVS \cite{li2017cifar10}, and N-Caltech101 \cite{orchard2015converting} datasets. We use direct coding to convert float pixel values into binary spikes \cite{wu2019direct}. In all our experiments, the spiking transformer STAtten models are trained from scratch. The detailed training strategy for each dataset is explained in the Supplementary Material. 

\noindent\textbf{Notation:} In the experiments, we use the notation STAtten + [$\cdot$] to denote STAtten implementation based on the backbone architecture used in [$\cdot$]. E.g., STAtten+\cite{zhou2022spikformer} refers to training a STAtten model with Spikformer backbone from scratch.

\begin{table*}[t]
\footnotesize
  \caption{Performance comparison between our methods and previous works on ImageNet datasets. Note that $*$ represents the inference accuracy with 288 $\times$ 288 resolution. In the architecture column, $L$-$D$ represents $L$ number of encoder blocks and $D$ hidden dimensions. $\dagger$ represents our implementation.}
  \label{imagenet_result}
  \centering
  \begin{tabular}{llccccc}
    \toprule
    Method & Type &Architecture  & Param(M) & Timestep & Accuracy($\%$)   \\
    \midrule
     Vision Transformer \cite{dosovitskiy2020image}& ANN & ViT-B/16 & 86.59 & 1 & 77.90 \\
     DeiT \cite{touvron2021training}& ANN & DeiT-B & 86.59 & 1 & 81.80 \\
     Swin Transformer \cite{liu2021swin}& ANN & Swin Transformer-B & 87.77 & 1 & 83.50 \\
     \cdashline{1-6}\\[-1.75ex]
     TET \cite{deng2022temporal}& SNN & SEW-ResNet34 & 21.79 & 4 & 68.00 \\
     Spiking ResNet \cite{hu2021spiking}& SNN & ResNet50 & 25.56 & 350 &72.75  \\
     tdBN \cite{zheng2021going}& SNN & ResNet34 & 21.79 & 6 & 63.72 \\
     SEW-ResNet \cite{fang2021deep}& SNN & SEW-ResNet152 & 60.19 & 4 & 69.26 \\
     MS-ResNet \cite{hu2024advancing}& SNN & ResNet104 & 78.37 & 5 & 74.21 / 76.02* \\
     Att-MS-ResNet \cite{yao2023attention}& SNN & ResNet104 & 78.37 & 4 & 77.08*  \\
     \cdashline{1-6}\\[-1.75ex]
     
     Spikformer \cite{zhou2022spikformer}& SNN& Spikformer-8-768  & 66.34 & 4 & 74.81 \\
     SpikingReformer \cite{shi2024spikingresformer}& SNN & SpikingReformer-6-1024 & 66.38 & 4 & 78.77 / 79.40* \\
     QKFormer \cite{zhou2024qkformer} & SNN& QKFormer-6-1024 & 64.96 & 4 & 84.22 \\
     
     \midrule
     \multirow{3}{*}{Spike-driven Transformer \cite{yao2024spike}}& SNN  & Spike-driven Transformer-8-512 & 29.68 & 4 & 74.57 \\
     & SNN& Spike-driven Transformer-10-512 & 36.01 & 4 & 74.66 \\
     & SNN& Spike-driven Transformer-8-768 & 66.34 & 4 & 76.32 / 77.07* \\
    \cdashline{1-6}\\[-1.75ex]
    
     \cellcolor{gray!25}& \cellcolor{gray!25}SNN& \cellcolor{gray!25}Spike-driven Transformer-8-512 & \cellcolor{gray!25}29.68 & \cellcolor{gray!25}4 & \cellcolor{gray!25}76.18 / 76.56* \\
          \multirow{-2}{*}{\cellcolor{gray!25}\textbf{STAtten} + \cite{yao2024spike}}  & \cellcolor{gray!25}SNN& \cellcolor{gray!25}Spike-driven Transformer-8-768 & \cellcolor{gray!25}66.34 & \cellcolor{gray!25}4 & \cellcolor{gray!25}\textbf{78.11} / \textbf{78.39*} \\
    \midrule
    Spike-driven Transformer-V2 \cite{yao2024spike2}& SNN & Spike-driven Transformer-V2-8-512 & 55.4 & 4 & 79.49$\dagger$ / 79.98*$\dagger$ \\
    \cdashline{1-6}\\[-1.75ex]
    \cellcolor{gray!25}\textbf{STAtten} + \cite{yao2024spike2}& \cellcolor{gray!25}SNN & \cellcolor{gray!25}Spike-driven Transformer-V2-8-512 & \cellcolor{gray!25}55.4 & \cellcolor{gray!25}4 & \cellcolor{gray!25}\textbf{79.85 / 80.67*} \\
    \bottomrule
  \end{tabular}
\end{table*}

\begin{table*}[t]
\footnotesize
  \caption{Performance comparison between our methods and previous works on CIFAR10-DVS and N-Caltech101 datasets. In the architecture column, $L$-$D$ represents $L$ number of encoder blocks and $D$ hidden dimensions. All baseline architectures of spike-based transformer and corresponding STAtten implementations are trained from scratch for fair comparison.}
  \label{dvs_result}
  \centering
  \begin{tabular}{llcccc}
    \toprule
    Method & Architecture  & Timestep & CIFAR10-DVS ($\%$) & N-Caltech101 ($\%$)      \\
    \midrule
     tdBN \cite{zheng2021going} & ResNet19 & 10 & 67.80 & - \\
     PLIF \cite{fang2021incorporating} & ConvNet & 20 & 74.80 & - \\
     Dspike \cite{li2021differentiable} & ResNet18 & 10 & 75.40 & - \\
     DSR \cite{meng2022training} & ResNet18 & 20 & 77.27 & - \\
     SEW-ResNet \cite{fang2021deep} & ConvNet & 20 & 74.80 & - \\
     TT-SNN \cite{lee2024tt} & ResNet34 & 6 & - & 77.80 \\
     NDA \cite{li2022neuromorphic} & VGG11 & 10 & 79.6 & 78.2 \\
     
     Spikformer \cite{zhou2022spikformer} & Spikformer-2-256 & 16 & 80.9 & - \\
     \midrule
     Spike-driven Transformer \cite{yao2024spike}  & Spike-driven Transformer-2-256 & 16 & 80.0 & 81.80 \\
     \cdashline{1-5}\\[-1.75ex]
     \cellcolor{gray!25}\textbf{STAtten} + \cite{yao2024spike} &  \cellcolor{gray!25}Spike-driven Transformer-2-256 & \cellcolor{gray!25}16& \cellcolor{gray!25}\textbf{81.1}& \cellcolor{gray!25} \textbf{83.15} \\
     \midrule
     SpikingReformer \cite{shi2024spikingresformer}   & SpikingReformer-4-384 & 16 & 78.80 & 81.29  \\
     \cdashline{1-5}\\[-1.75ex]
     \cellcolor{gray!25}\textbf{STAtten} + \cite{shi2024spikingresformer}&  \cellcolor{gray!25}SpikingReformer-4-384 & \cellcolor{gray!25}16& \cellcolor{gray!25}\textbf{80.60}& \cellcolor{gray!25}\textbf{81.95} \\
     \midrule
     QKFormer \cite{zhou2024qkformer}  & QKFormer-4-384 & 16 & 82.90 & 83.58  \\
     \cdashline{1-5}\\[-1.75ex]
     \cellcolor{gray!25}\textbf{STAtten} + \cite{zhou2024qkformer}&  \cellcolor{gray!25}SpikingReformer-4-384 & \cellcolor{gray!25}16& \cellcolor{gray!25}\textbf{83.90}& \cellcolor{gray!25}\textbf{84.25} \\
     
    \bottomrule
    \vspace{-8mm}
  \end{tabular}
\end{table*}

\subsection{Sequential CIFAR10/100 Classification}

We first evaluate our spatial-temporal attention mechanism's capability to capture long-term temporal dependencies using sequential CIFAR10 and CIFAR100 datasets. In this task, we transform the standard image classification task into a temporal sequence processing task by dividing each input image column-wise, shown in Fig. \ref{heatmap}, where each column serves as input for one timestep. This results in a sequence length of 32 timesteps, corresponding to the image width. To accommodate this 1-D sequential input, we modify the convolutional, batch normalization, and pooling layers from 2-D to 1-D operations. As shown in Table \ref{seq_tab}, STAtten outperforms existing spatial-only attention methods on both datasets. Specifically, in Spike-driven Transformer \cite{yao2024spike}, our approach achieves improvements of 3.09 \% and 3.22 \% on s-CIFAR10 and s-CIFAR100, respectively, compared to the baseline models. This performance gain demonstrates STAtten's capability in capturing long-range temporal dependencies.





\vspace{-2mm}
\subsection{Performance Analysis}
\vspace{-1mm}

\textbf{CIFAR10/100. } The CIFAR10/100 datasets comprise static images of 32 $\times$ 32 pixels, with 50,000 training and 10,000 test images. As shown in Table \ref{cifar_result}, STAtten improves the performance of various spike-based transformers, achieving 95.35\% accuracy on CIFAR10 and 80.20\% on CIFAR100 when integrated with QKFormer. These improvements over spatial-only attention methods are consistent across different architectures, demonstrating the benefits of spatial-temporal processing.

\noindent\textbf{ImageNet.} 
ImageNet is a large-scale image dataset with images sized 224 $\times$ 224 pixels, comprising approximately 1.2 million training images and 50,000 validation images across 1,000 classes. As shown in Table \ref{imagenet_result}, STAtten enhances the performance of spike-based transformers. With Spike-driven Transformer \cite{yao2024spike}, it achieves 78.11\% accuracy at 224 $\times$ 224 resolution and 78.39\% at 288 $\times$ 288 resolution. Integration with Spike-driven Transformer-V2 \cite{yao2024spike2} further improves performance to 79.85\% and 80.67\% at respective resolutions. It is noteworthy that Spike-driven Transformer \cite{yao2024spike} achieves comparable or better performance than recent Spikformer models while using only 29.68M parameters. Furthermore, STAtten with Spike-driven Transformer-V2 achieves even higher accuracy than SpikingReformer and Spikformer while using around 20\% fewer parameters. 



\noindent\textbf{Neuromorphic Datasets. } We evaluate our approach on two neuromorphic vision datasets: CIFAR10-DVS \cite{li2017cifar10} and N-Caltech101 \cite{orchard2015converting}. CIFAR10-DVS contains 10,000 samples (9,000 training, 1,000 testing) captured using a Dynamic Vision Sensor (DVS) from static CIFAR10 images. N-Caltech101, derived from the Caltech101 dataset \cite{li2022caltech}, comprises 8,831 DVS image samples. Following standard protocols, we resize CIFAR10-DVS samples to 64 $\times$ 64 and apply the augmentation strategy from \cite{yao2024spike}, while N-Caltech101 samples are resized to 64 $\times$ 64 with NDA \cite{li2022neuromorphic} augmentation. As shown in Table \ref{dvs_result}, integrating STAtten consistently improves performance across different architectures. With Spike-driven Transformer as the backbone, our approach achieves 81.1\% accuracy on CIFAR10-DVS and 83.15\% on N-Caltech101, improving the baseline by 1.1\% and 1.35\% respectively. The enhancement extends to other architectures as well, with SpikingReformer achieving 80.60\% on CIFAR10-DVS and 81.95\% on N-Caltech101, and QKFormer reaching 83.90\% and 84.25\% respectively, which is the state-of-the-art result.


\vspace{-2mm}
\subsection{Memory and Energy Analysis}
\label{main_energy}
We analyze memory and energy consumption during ImageNet inference. As shown in Table \ref{energy_tab}, traditional transformer, ViT-B/16, with 32-bit precision requires high memory (351.8 MB) and energy consumption (254.84 mJ). Even with 8-bit precision, Liu~\cite{liu2021post} still consumes 110.8 mJ of energy. In contrast, spike-based transformers with 32-bit weights and binary activations demonstrate significant energy efficiency. Notably, STAtten integrated with SDT and SDT-V2 achieves similar energy consumption to the baseline architectures, showing that our spatial-temporal processing can be realized without additional energy overhead. Correspondingly, STAtten attains comparable performance with fewer parameters, as detailed in the next section. The methodology for energy calculations is provided in the Supplementary Material.

\begin{table}
\centering
\caption{Comparison of memory, energy, and accuracy on ImageNet across different transformer architectures. The precision format weight/activation bit-width and the memory include the size of model parameters and activations. We show the memory consumption for both 32-bit and 8-bit weight spiking models. SDT represents Spike-driven Transformer.}
\vspace{-2mm}
\label{energy_tab}
\resizebox{\linewidth}{!}{%
\begin{tabular}[width=0.85\textwidth]{lcccc}
    \toprule
    Method &Precision & Timestep & Memory (MB) & Energy (mJ)   \\
    \midrule
    ViT-B/16 \cite{dosovitskiy2020image} & 32/32 & 1 & 351.8 &  254.84  \\
    \cdashline{1-5}\\[-1.75ex]
    Liu (ViT-B) \cite{liu2021post}  & 8/8 & 1 & 87.96 & 110.8 \\
    \midrule
    Spikformer \cite{zhou2022spikformer}&32(8)/1 & 4 & 285.99 (86.48) & 21.48  \\
     SpikingReformer \cite{shi2024spikingresformer} & 32(8)/1 & 4 & 273.57 (85.4) & 8.76  \\
     QKFormer \cite{zhou2024qkformer} & 32(8)/1 & 4 & 280.28 (84.99) & 38.91  \\
     \cdashline{1-5}\\[-1.75ex]
     SDT \cite{yao2024spike}&32(8)/1 & 4 & 283.97 (87.42) & 12.42 \\ 
     STAtten + \cite{yao2024spike}&32(8)/1& 4 &283.97 (87.42) & 12.36  \\
     \cdashline{1-5}\\[-1.75ex]
     SDT-V2 \cite{yao2024spike2}&32(8)/1 & 4 & 250.22 (84.02)& 52.40  \\ 
     STAtten + \cite{yao2024spike}&32(8)/1& 4 & 250.22 (84.02)& 52.38  \\
    
    \bottomrule
    \end{tabular}}
\end{table}

\begin{figure}[t]
\centering
\begin{tabular}{@{}c@{\hskip 0.01\linewidth}c@{\hskip 0.01\linewidth}c}
\includegraphics[width=0.49\linewidth]{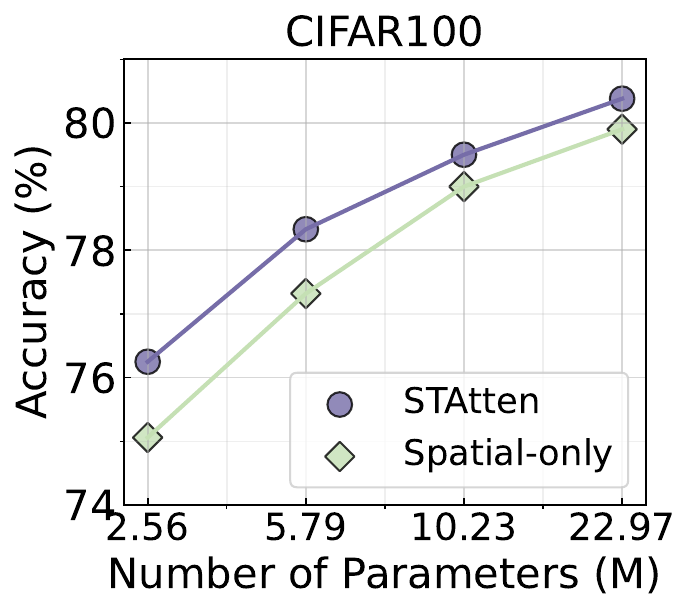} &
\includegraphics[width=0.49\linewidth]{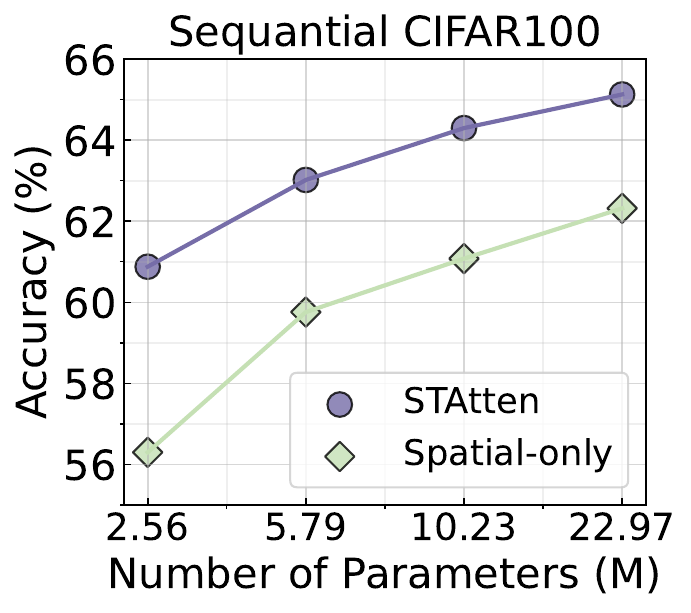}
\\
\vspace{-2mm}
{(a)} & {(b)} \\
\end{tabular}
\caption{Accuracy comparison with respect to the number of parameters on (a) CIFAR100 and (b) Sequential CIFAR100 datasets. Spike-driven Transformer \cite{yao2024spike} is used as the baseline of spatial-only architecture.}
\vspace{-7mm}
\label{fig:capacity}
\end{figure}

\vspace{-1mm}
\subsection{Model Capacity}
To further evaluate the capacity of our STAtten in spike-based transformers, we compare the accuracy based on the number of trainable parameters through ablation studies. Fig. \ref{fig:capacity} shows the performance scaling across different model sizes on both (a) standard CIFAR100 and (b) sequential CIFAR100 datasets. Spike-driven Transformer \cite{yao2024spike} is used as the baseline of spatial-only architecture. As shown in Fig. \ref{fig:capacity}(a), on CIFAR100, STAtten demonstrates strong performance across various model sizes ranging from 2.56M to 22.97M parameters. The accuracy improvement ($\sim0.5\text{-}1.0\%$) is consistently maintained across different architectural scales. For Sequential CIFAR100 (Fig. \ref{fig:capacity}(b)), where temporal dependencies are more crucial, STAtten shows particularly robust performance improvement ($\sim3\text{-}5\%$)  even with smaller architectures. 




\vspace{-1mm}
\subsection{Limitation}
The hardware implementation of STAtten faces challenges despite its improved accuracy. Full spatial-temporal attention without block partitioning is impractical for neuromorphic deployment, as it requires complete temporal information. Our block-wise approach partially addresses this by processing temporal information in steps, where each step corresponds to one block size. However, even with this optimization, deployment remains challenging on traditional neuromorphic chips like TrueNorth \cite{akopyan2015truenorth} and Loihi \cite{davies2018loihi} that process information step-by-step. A potential solution lies in layer-by-layer neuromorphic chips \cite{zhou20240, kim2023c, yin2022sata, yin2024loas} that support layer-wise processing across timesteps. This architectural shift provides an efficient implementation path for STAtten's block-wise processing. Future integration of parallel LIF neurons~\cite{fang2024parallel} could further accelerate computation in layer-by-layer architectures.


\vspace{-2mm}
\section{Conclusion}
This paper proposes a block-wise spatial-temporal attention mechanism, STAtten, that enhances spike-based transformers. Through block-wise processing and leveraging non-softmax properties, STAtten captures spatial-temporal information without additional overhead. STAtten can be integrated into existing architectures while preserving their energy efficiency, consistently improving performance across different backbones. Experimental results across both static and neuromorphic datasets validate the effectiveness of our approach.

\vspace{-2mm}
\section*{Acknowledgment} This work was supported in part by CoCoSys, a JUMP2.0 center sponsored by DARPA and SRC, the National Science Foundation (CAREER Award, Grant \#2312366, Grant \#2318152), the DARPA Young Faculty Award and the DoE MMICC center SEA-CROGS (Award \#DE-SC0023198).

{
    \newpage
    \small
    \bibliographystyle{ieeenat_fullname}
    \bibliography{main}
}

\clearpage
\setcounter{page}{1}
\maketitlesupplementary
\appendix


\section{Energy Calculation Details}

\label{appendix_energy}
To clarify the energy consumption of our STAtten architecture in Section \ref{main_energy}, we present the detailed equations of every layer shown in Table \ref{table_energy}. 
\begin{table}[h]
\footnotesize
  \caption{The detailed equations of energy consumption on every layer of STAtten architecture.}
  \label{table_energy}
  \centering
  \begin{tabular}{ccc}
    \toprule
    Block& Layer &Energy Consumption  \\
    \midrule
      \multirow{2}{*}{Embedding} & First Conv &  $E_{MAC} \cdot F_{Conv} \cdot T$  \\
                                 & Other Convs &  $E_{AC} \cdot F_{Conv} \cdot T \cdot S_{Conv}$ \\
    \cdashline{1-3}\\[-1.75ex]
     \multirow{3}{*}{Attention} & $Q$, $K$, $V$ &  $3 \cdot E_{AC} \cdot F_{Conv} \cdot T \cdot S_{Conv}$  \\
                            & Self-attention &  $E_{AC} \cdot TND^2 \cdot (S_K + S_V + S_Q)$ \\
                            & MLP &  $E_{AC} \cdot F_{Conv} \cdot T \cdot S_{Conv}$ \\

    \cdashline{1-3}\\[-1.75ex]
    \multirow{2}{*}{MLP} & MLP1 &  $E_{AC} \cdot F_{Conv} \cdot T \cdot S_{Conv}$  \\
                          & MLP2 &  $E_{AC} \cdot F_{Conv} \cdot T \cdot S_{Conv}$ \\
                                 
    \bottomrule
  \end{tabular}
\end{table}

Here, $E_{MAC}$ is the energy of MAC operation, $F_{Conv}$ is FLOPs of the convolutional layer, $T$, $N$, and $D$ are timestep, the number of patches, and channel dimension respectively, $S_{Conv}$ is the firing rate of input spikes on the convolutional layer, $S_Q$, $S_K$, and $S_V$ are the firing rate of input spikes on $Q$, $K$, and $V$ projection layer respectively. The FLOPs of the convolutional layer can be calculated as follows:
\begin{equation}
    F_{Conv} = K \cdot K \cdot H_{out} \cdot W_{out} \cdot C_{in} \cdot C_{out},
\end{equation}
where $K$ is kernel size, $H_{out}$ and $W_{out}$ are the height and width of the output feature map respectively, and $C_{in}$ and $C_{out}$ are the input and output channel dimension respectively.

In the embedding block, for the first convolutional layer, since we use direct coding to convert a float pixel value into binary spikes \cite{wu2019direct}, the firing rate does not need to be calculated for energy consumption, and $E_{MAC}$ is used for the float pixel input. In the Attention block, for the energy calculation of the self-attention part, we can use the equations of our spatial-temporal methods shown in Table \ref{table1}. Following previous works \cite{zhou2022spikformer, yao2024spike}, we calculate the energy consumption based on the FLOPs operation executed in 45nm CMOS technology \cite{horowitz20141}, \eg, $E_{MAC}=4.6pJ$, and $E_{AC}=0.9pJ$. The firing rate and the theoretical energy consumption of the STAtten with Spike-driven Transformer architecture are provided in Appendix \ref{fire}.

\section{Experimental Details}
\label{appendix_ex}
In this section, we provide the experimental details on CIFAR10/100, ImageNet, CIFAR10-DVS, and N-Caltech101 datasets. The Table \ref{appendix_exp_table} shows that general experimental setup in \cite{yao2024spike}. In other architecture \cite{yao2024spike2, zhou2022spikformer, shi2024spikingresformer, zhou2024qkformer}, we follow their configurations for fair comparison.

\begin{table}
\footnotesize
  \caption{The experimental details on each dataset. $L$-$D$ in architecture represents $L$ number of encoder blocks and $D$ channel dimension.}
  \label{appendix_exp_table}
  \centering
  \begin{tabular}{ccccc}
    \toprule
    & CIFAR10/100 & ImageNet& DVS  \\
    \midrule
    Timestep & 4 &  4 & 16 \\
    Batch size & 64 &  32 & 16\\ 
    Learning rate & 0.0003 &  0.001 & 0.01\\ 
    Training epoch & 310 &  210 & 210\\ 
    Optimizer & AdamW &  Lamb & AdamW \\
    Hardware (GPU) & A5000 &  A100 & A5000\\ 
    Architecture & 2-512 &  8-768 & 2-256 \\
    \bottomrule
  \end{tabular}
\end{table}

We apply data augmentation following \cite{zhou2022spikformer, yao2024spike}. For the ImageNet dataset, general augmentation techniques such as random augmentation, mixup, and cutmix are employed. Different data augmentation strategies are applied to the CIFAR10-DVS and N-Caltech101 datasets according to NDA \cite{li2022neuromorphic}. While training on the dynamic datasets, we add a pooling layer branch and a residual connection to the spatial-temporal attention layer. The outputs of the pooling layer and the spatial-temporal attention are then multiplied element-wise to extract important spike feature maps.

\begin{table*}[t]

\centering
\caption{Analysis of temporal block combinations and their accuracy. Each entry shows timestep ranges for Q/K/V tensors across two blocks ($B_1$, $B_2$). For example, [1,2]/[3,4]/[1,2] indicates Q and V use timesteps 1-2 while K uses timesteps 3-4. Notation [0:16] represents timesteps from 0 through 16.}
\label{combination}
\begin{tabular}{cccc}

\toprule
\multirow{2}{*}{Dataset}& \multicolumn{2}{c}{Temporal Combination ($\mathbf{Q}/\mathbf{K}/\mathbf{V}$)}& \multirow{2}{*}{Accuracy (\%)} \\
& $B_1$& $B_2$&\\
\midrule
\multirow{3}{*}{\begin{tabular}[c]{@{}c@{}}CIFAR100\end{tabular}} & {[}1,2{]} / {[}1,2{]} / {[}1,2{]}   & {[}3,4{]} / {[}3,4{]} / {[}3,4{]}    & 79.85\\
\cdashline{2-4}\\[-1.75ex]
& {[}1,2{]} / {[}3,4{]} / {[}1,2{]}    & {[}3,4{]} / {[}1,2{]} / {[}3,4{]}  &79.28\\
\cdashline{2-4}\\[-1.75ex]
& {[}1,4{]} / {[}2,3{]} / {[}1,4{]}    & {[}2,3{]} / {[}1,4{]} / {[}2,3{]} &79.09\\
\midrule
\multirow{2}{*}{\begin{tabular}[c]{@{}c@{}}Sequential CIFAR100\end{tabular}}& {[}0:16{]} / {[}0:16{]} / {[}0:16{]} & {[}16:32{]} / {[}16:32{]} / {[}16:32{]} & 62.95\\
\cdashline{2-4}\\[-1.75ex]
& {[}0:16{]} / {[}16:32{]} / {[}0:16{]} & {[}16:32{]} / {[}0:16{]} / {[}16:32{]} &62.80\\
\midrule
\multirow{2}{*}{\begin{tabular}[c]{@{}c@{}}N-Caltech101\end{tabular}}& {[}0:8{]} / {[}0:8{]} / {[}0:8{]}  & {[}8:16{]} / {[}8:16{]} / {[}8:16{]} &82.49\\
\cdashline{2-4}\\[-1.75ex]
& {[}0:8{]} / {[}8:16{]} / {[}0:8{]}    & {[}8:16{]} / {[}0:8{]} / {[}8:16{]}     & 79.09\\
\bottomrule
    
\end{tabular}
\end{table*}

\section{Ablation Study}
In this section, we analyze the impact of timestep combinations and block sizes in our block-wise attention mechanism.

\subsection{Timestep Combination}

In section \ref{motivation}, we identified that binary matrix multiplication between temporally distant spikes can increase silent neurons, leading to information loss. This phenomenon can be explained through binary matrix multiplication patterns. Let $\mathbf{Q}_t, \mathbf{K}_{t'} \in \{0,1\}^{N \times D}$ be binary spike matrices at timesteps $t$ and $t'$. When computing attention between these timesteps, each element of their product is:
\begin{equation}
 (\mathbf{Q}_t\mathbf{K}_{t'}^\top)_{i,j} = \sum_{d=1}^D q_{t,i,d} \cdot k_{t',j,d},
\end{equation}
where $i,j \in \{1,...,N\}$ represent token positions, and $d \in \{1,...,D\}$ is the feature dimension. As the temporal distance $|t-t'|$ increases, the spike patterns become less correlated, increasing the probability of $q_{t,i,d} \cdot k_{t',j,d} = 0$. This multiplicative effect accumulates across the dimension $D$, leading to more zero outputs and consequently more silent neurons.

To illustrate this effect, consider binary matrices $\mathbf{Q}_t$ and $\mathbf{K}_{t'}$ with the same number of spikes but at different temporal distances. For nearby timesteps $t$ and $t+1$:
\begin{equation}
\mathbf{Q}_t = \begin{bmatrix} 
1 & 1 & 0 & 1 & 0 \\
0 & 1 & 1 & 0 & 1 \\
1 & 0 & 1 & 1 & 0 \\
1 & 0 & 1 & 0 & 1
\end{bmatrix}, 
\mathbf{K}_{t+1} = \begin{bmatrix} 
1 & 1 & 0 & 1 & 0 \\
0 & 1 & 1 & 0 & 1 \\
1 & 0 & 1 & 0 & 1 \\
1 & 1 & 0 & 1 & 0
\end{bmatrix}
\end{equation}
Their product yields many high values due to similar patterns:
\begin{equation}
\mathbf{Q}_t\mathbf{K}_{t+1}^\top = \begin{bmatrix} 
3 & 2 & 2 & 3 \\
2 & 3 & 2 & 1 \\
2 & 2 & 2 & 2 \\
2 & 2 & 2 & 2
\end{bmatrix}
\end{equation}
However, for distant timesteps $t$ and $t+\Delta$:
\begin{equation}
\mathbf{Q}_t = \begin{bmatrix} 
1 & 1 & 0 & 1 & 0 \\
0 & 1 & 1 & 0 & 1 \\
1 & 0 & 1 & 1 & 0 \\
1 & 0 & 1 & 0 & 1
\end{bmatrix}, 
\mathbf{K}_{t+\Delta} = \begin{bmatrix} 
0 & 1 & 1 & 0 & 1 \\
1 & 0 & 1 & 1 & 0 \\
1 & 1 & 0 & 0 & 1 \\
0 & 1 & 1 & 1 & 0
\end{bmatrix}
\end{equation}
Their product contains low values and zeros despite having the same spike density:
\begin{equation}
\mathbf{Q}_t\mathbf{K}_{t+\Delta}^\top = \begin{bmatrix} 
1 & 1 & 1 & 1 \\
1 & 1 & 0 & 1 \\
1 & 1 & 1 & 1 \\
0 & 1 & 1 & 1
\end{bmatrix}
\end{equation}

Since we apply LIF after $\mathbf{Q}\mathbf{K}^\top\mathbf{V}$ operations to generate spikes, matrices with higher values from nearby timesteps are more likely to trigger neurons compared to lower values from distant timesteps. This example demonstrates that temporal distance leads to less correlated spike patterns, resulting in increased silent neurons. Fig.~\ref{reason}(b) visualizes this effect on the CIFAR100 dataset, showing higher neuron activation when correlating nearby timesteps compared to distant ones.

Table \ref{combination} shows the performance comparison across different datasets by varying temporal combinations of $\mathbf{Q}$, $\mathbf{K}$, and $\mathbf{V}$. The notation [a,b]/[c,d]/[e,f] indicates that $\mathbf{Q}$, $\mathbf{K}$, and $\mathbf{V}$ use timesteps [a,b], [c,d], and [e,f], respectively. For instance, in CIFAR100's $B_1$, [1,2]/[3,4]/[1,2] means $\mathbf{Q}$ and $\mathbf{V}$ use timesteps 1-2 while $\mathbf{K}$ uses timesteps 3-4. Across all datasets, combinations using different timestep ranges consistently show lower performance compared to those using the same ranges.

\begin{table}[]

\centering
\caption{Accuracy comparison with different block sizes. $T$ represents the timestep for each dataset, $B$ denotes the block size.}
\label{blocksize}
\begin{tabular}{cccc}

\toprule
Dataset& Block size& Accuracy (\%) \\
\midrule
\multirow{2}{*}{\begin{tabular}[c]{@{}c@{}}CIFAR100\\($T=4$)\end{tabular}} & B=2 & 79.85\\
\cdashline{2-4}\\[-1.75ex]
&B=4&79.90\\

\midrule
\multirow{3}{*}{\begin{tabular}[c]{@{}c@{}}ImageNet\\($T=4$)\end{tabular}}& B=1 & 77.65\\
\cdashline{2-4}\\[-1.75ex]
& B=2 &78.00\\
\cdashline{2-4}\\[-1.75ex]
& B=4 &78.06\\

\midrule
\multirow{3}{*}{\begin{tabular}[c]{@{}c@{}}Sequential CIFAR100\\($T=32$)\end{tabular}}& B=8 &60.89\\
\cdashline{2-4}\\[-1.75ex]
& B=16 &62.95\\
\cdashline{2-4}\\[-1.75ex]
& B=32 & 64.30\\
\midrule
\multirow{3}{*}{\begin{tabular}[c]{@{}c@{}}N-Caltech101\\($T=16$)\end{tabular}}& B=4&  83.15\\
\cdashline{2-4}\\[-1.75ex]
& B=8&  82.49\\
\cdashline{2-4}\\[-1.75ex]
& B=16 &82.40\\
\bottomrule
    
\end{tabular}
\end{table}

\subsection{Block Size Analysis}
STAtten employs block-wise processing for memory efficiency. Table \ref{blocksize} shows how block size affects performance across different datasets. For CIFAR100 ($T$=4), using block size $B$=2 achieves comparable accuracy to full spatial-temporal correlation ($B$=4), with only 0.05\% difference. Similarly, for ImageNet ($T$=4), block sizes $B$=2, and $B$=4 yield accuracies of 78.00\%, and 78.06\%, indicating that larger block sizes slightly improve performance. However, sequential CIFAR100 ($T$=32) shows an opposite trend: smaller block sizes lead to decreased accuracy because temporal information dominates spatial features in this dataset. Therefore, we use B=32 for the results presented in Table \ref{seq_tab}. For N-Caltech101 ($T$=16), we achieve optimal performance with $B$=4. This reveals that optimal block size depends on temporal-to-spatial information ratio: vision tasks favor smaller blocks to preserve spike correlation, while sequential tasks need larger blocks for temporal modeling.

\section{Versatility in Vision Tasks}
To demonstrate the generalizability and robustness of our STAtten, we extend its application to additional vision tasks, including object detection and transfer learning. 

\subsection{Object Detection}
We evaluate the adaptability of STAtten in the object detection domain by integrating it as the backbone in the EMS-YOLO~\cite{su2023deep} framework, replacing the original backbone. We train the model on the PASCAL VOC dataset~\cite{everingham2010pascal} from scratch, maintaining the same training configuration as the baseline~\cite{su2023deep} for a fair comparison. The results, presented in Table~\ref{tab:statten_detection}, demonstrate STAtten’s competitive performance in object detection compared to other spike-based architectures~\cite{kim2020spiking, yao2024spike}. These results highlight STAtten’s adaptability to diverse vision tasks beyond classification.

\subsection{Transfer Learning}
To further validate the generalizability of STAtten, we conduct transfer learning experiments on CIFAR-10 and CIFAR-100 datasets. We leverage pre-trained weights from ImageNet and resize the input images to 224$\times$224 pixels to align with standard transfer learning protocols. The results, also shown in Table~\ref{tab:statten_transfer}, indicate that STAtten achieves top performance in transfer learning tasks. These results underscore STAtten’s ability to generalize effectively across datasets and tasks, leveraging its spatial-temporal attention mechanism to extract robust features from pre-trained weights.

\begin{table}[ht]
\footnotesize
  \centering
  \caption{Performance comparison between STAtten and previous works on object detection using PASCAL VOC dataset.}
  \label{tab:statten_detection}
  \begin{tabular}{lcc}
    \toprule
    Method & mAP@0.5 (\%) & mAP@0.5:0.9 (\%) \\
    \midrule
    Spiking-YOLO~\cite{kim2020spiking}& 51.83 & - \\
    SDT~\cite{yao2024spike} & 51.63 & 25.31 \\
    \cdashline{1-3}\\[-1.75ex]
    \textbf{STAtten + \cite{yao2024spike}} & \textbf{52.98} & \textbf{27.53} \\
    \bottomrule
  \end{tabular}
\end{table}

\begin{table}[ht]
\footnotesize
  \centering
  \caption{Performance comparison between STAtten and previous works on transfer learning using ImageNet pre-trained weights on CIFAR-10 and CIFAR-100.}
  \label{tab:statten_transfer}
  \begin{tabular}{lcc}
    \toprule
    Method & CIFAR-10 (\%) & CIFAR-100 (\%) \\
    \midrule
    Spikformer~\cite{zhou2022spikformer} & 97.03 & 83.83 \\
    SpikingReformer~\cite{shi2024spikingresformer} & 97.40 & 85.98 \\
    \cdashline{1-3}\\[-1.75ex]
    \textbf{STAtten + \cite{yao2024spike}} & \textbf{97.76} & \textbf{86.67} \\
    \bottomrule
  \end{tabular}
\end{table}

\section{Firing rate}
\label{fire}
In this section, we present the firing rate and energy consumption of each layer in Spike-driven Transformer 8-768 architecture with STAtten, pre-trained with the ImageNet dataset. Note that the firing rates represent the firing rate of input spikes for each layer. Additionally, for the firing rate of Self-attention in the table below, we calculate it using the equation: $S_K + S_V + S_Q$.

\begin{table*}
\label{firing_rate}
  \centering
  \begin{NiceTabular}{cc|cccc|c}
    \toprule
    Block & Layer & $T=1$& $T=2$ & $T=3$ & $T=4$ & Energy ($mJ$) \\
    \midrule
    \multirow{5}{*}{Embedding} & 1st Conv & - & - & - & - & 0.5982\\
                               & 2nd Conv & 0.0771 & 0.1389 & 0.1092 & 0.1561 & 0.9015\\
                               & 3rd Conv & 0.0424 & 0.0644 & 0.0586 & 0.0527 & 0.4089\\
                               & 4th Conv & 0.0328 & 0.0501 & 0.0428 & 0.0480 & 0.3253\\
                               & 5th Conv & 0.0660 & 0.1402 & 0.1308 & 0.1413 & 0.4478\\
    \midrule
    \multirow{3}{*}{Encoder-1} & $Q$, $K$, $V$ & 0.2159 & 0.2662 & 0.2609 & 0.2728 & 0.3171\\
                               & Self-attention & 0.1221 & 0.1313 & 0.1320 & 0.1451 & 0.0993\\
                               & MLP & 0.2018 & 0.2962 & 0.2880 & 0.3454 & 0.1177\\
    \cdashline{1-7}\\[-1.75ex]
    \multirow{2}{*}{MLP-1} & MLP1 & 0.3292 & 0.3605 & 0.3622 & 0.3697 & 0.5916\\
                           & MLP2 & 0.0340 & 0.0409 & 0.0401 & 0.0458 & 0.0670\\

    \midrule
    \multirow{3}{*}{Encoder-2} & $Q$, $K$, $V$ & 0.3268 & 0.3583 & 0.3543 & 0.3967 & 0.4482\\
                               & Self-attention & 0.0986 & 0.0950 & 0.0945 & 0.1017 & 0.0867\\
                               & MLP & 0.2760 & 0.3532 & 0.3371 & 0.3511 & 0.1370\\
    \cdashline{1-7}\\[-1.75ex]
    \multirow{2}{*}{MLP-2} & MLP1 & 0.3094 & 0.3332 & 0.3321 & 0.3718 & 0.5604\\
                           & MLP2 & 0.0226 & 0.0293 & 0.0301 & 0.0350 & 0.0487\\ 

    \midrule
    \multirow{3}{*}{Encoder-3} & $Q$, $K$, $V$ & 0.3240 & 0.3462 & 0.3504 & 0.3917 & 0.4408\\
                               & Self-attention & 0.0752 & 0.0694 & 0.0680 & 0.0654 & 0.0772\\
                               & MLP & 0.2837 & 0.3409 & 0.3254 & 0.2879 & 0.1288\\
    \cdashline{1-7}\\[-1.75ex]
    \multirow{2}{*}{MLP-3} & MLP1 & 0.3486 & 0.3519 & 0.3588 & 0.3957 & 0.6056\\
                           & MLP2 & 0.0186 & 0.0241 & 0.0245 & 0.0255 & 0.0386\\         

    \midrule
    \multirow{3}{*}{Encoder-4} & $Q$, $K$, $V$ & 0.3532 & 0.3570 & 0.3661 & 0.4015 & 0.4613\\
                               & Self-attention & 0.0716 & 0.0707 & 0.0704 & 0.0743 & 0.0749\\
                               & MLP & 0.2586 & 0.3299 & 0.3246 & 0.3203 & 0.1283\\
    \cdashline{1-7}\\[-1.75ex]
    \multirow{2}{*}{MLP-4} & MLP1 & 0.3591 & 0.3544 & 0.3633 & 0.3965 & 0.6132\\
                           & MLP2 & 0.0138 & 0.0177 & 0.0183 & 0.0188 & 0.0286\\        

    \midrule
    \multirow{3}{*}{Encoder-5} & $Q$, $K$, $V$ & 0.3599 & 0.3588 & 0.3688 & 0.3979 & 0.4637\\
                               & Self-attention & 0.0701 & 0.0619 & 0.0610 & 0.0631 & 0.0694\\
                               & MLP & 0.2695 & 0.2588 & 0.2469 & 0.2187 & 0.1034\\
    \cdashline{1-7}\\[-1.75ex]
    \multirow{2}{*}{MLP-5} & MLP1 & 0.3645 & 0.3579 & 0.3691 & 0.3979 & 0.6199\\
                           & MLP2 & 0.0098 & 0.0126 & 0.0128 & 0.0134 & 0.0202\\

    \midrule
    \multirow{3}{*}{Encoder-6} & $Q$, $K$, $V$ & 0.3737 & 0.3621 & 0.3706 & 0.3941 & 0.4684\\
                               & Self-attention & 0.0740 & 0.0581 & 0.0533 & 0.0496 & 0.0606\\
                               & MLP & 0.2071 & 0.2260 & 0.1896 & 0.1393 & 0.0793\\
    \cdashline{1-7}\\[-1.75ex]
    \multirow{2}{*}{MLP-6} & MLP1 & 0.3832 & 0.3665 & 0.3743 & 0.3963 & 0.6327\\
                           & MLP2 & 0.0108 & 0.0128 & 0.0119 & 0.0107 & 0.0193\\   

    \midrule
    \multirow{3}{*}{Encoder-7} & $Q$, $K$, $V$ & 0.3815 & 0.3665 & 0.3663 & 0.3826 & 0.4672\\
                               & Self-attention & 0.0746 & 0.0575 & 0.0538 & 0.0528 & 0.0615\\
                               & MLP & 0.1802 & 0.1670 & 0.1362 & 0.0972 & 0.0604\\
    \cdashline{1-7}\\[-1.75ex]
    \multirow{2}{*}{MLP-7} & MLP1 & 0.3773 & 0.3574 & 0.3549 & 0.3686 & 0.6069\\
                           & MLP2 & 0.0056 & 0.0068 & 0.0068 & 0.0063 & 0.0106\\   

    \midrule
    \multirow{3}{*}{Encoder-8} & $Q$, $K$, $V$ & 0.3772 & 0.3423 & 0.3471 & 0.3594 & 0.4452\\
                               & Self-attention & 0.1180 & 0.0853 & 0.0784 & 0.0728 & 0.0926\\
                               & MLP & 0.1383 & 0.1324 & 0.1143 & 0.1010 & 0.0505\\
    \cdashline{1-7}\\[-1.75ex]
    \multirow{2}{*}{MLP-8} & MLP1 & 0.3684 & 0.3480 & 0.3616 & 0.3818 & 0.6075\\
                           & MLP2 & 0.0123 & 0.0177 & 0.0168 & 0.0145 & 0.0255\\  
    \bottomrule
  \end{NiceTabular}
\end{table*}

\end{document}